\pdfoutput=1
\documentclass[10pt,twocolumn,letterpaper]{article}

\usepackage[pagenumbers]{cvpr}      %
\usepackage{xcolor}

\usepackage{amsmath,amsthm,amssymb,booktabs}

\DeclareMathOperator{\Cov}{Cov}
\DeclareMathOperator{\Std}{Std}
\providecommand{\E}{\mathbb{E}}

\AtBeginDocument{%
  \setlength{\abovedisplayskip}{5pt plus 1pt minus 2pt}%
  \setlength{\belowdisplayskip}{5pt plus 1pt minus 2pt}%
  \setlength{\abovedisplayshortskip}{2pt plus 1pt}%
  \setlength{\belowdisplayshortskip}{3pt plus 1pt minus 1pt}}

\usepackage{placeins}

\definecolor{cvprblue}{rgb}{0.21,0.49,0.74}
\usepackage[pagebackref,breaklinks,colorlinks,allcolors=cvprblue]{hyperref}

\title{Beyond Majority Vote: Judge-Guided Test-Time Reinforcement Learning\\for Vision-Language Models}

\author{Parsa Esmaeilkhani \qquad Longin Jan Latecki\\
Temple University\\
Philadelphia, USA\\
{\tt\small parsa.esmaeilkhani@temple.edu \qquad latecki@temple.edu}
}

\begin{document}
\twocolumn[{%
\renewcommand\twocolumn[1][]{#1}%
\maketitle
\begin{center}
\expandafter\def\csname @captype\endcsname{figure}
\includegraphics[width=\textwidth]{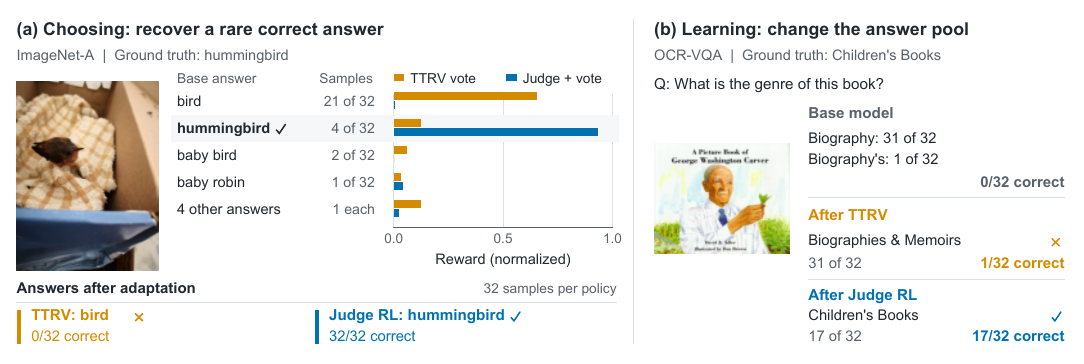}
\vspace{-4mm}
\caption{\textbf{A correct answer can lose the vote.} Test-time RL adapts a model using only unlabeled questions: the model samples several answers to each question, a reward scores them, and the model is updated toward the high-reward answers. Without labels, methods such as TTRV reward the answer most samples agree on. (a) On this ImageNet-A image, Qwen2.5-VL-7B says \emph{bird} in 21 of 32 samples and the correct \emph{hummingbird} in 4. The vote reward favors \emph{bird}; our reward adds a learned judge's score and favors \emph{hummingbird}. After adaptation, all 32 samples say \emph{bird} under TTRV and \emph{hummingbird} under ours. (b) Adaptation can produce a correct answer absent from the 32 base-model samples. On this OCR-VQA cover, 17 of 32 samples are correct after adaptation.}
\label{fig:teaser}
\end{center}
}]

\begin{abstract}
Test-time reinforcement learning (RL) adapts a vision-language model (VLM) to a new distribution using only unlabeled questions, typically by rewarding the answer that most of the model's samples agree on. This works on multiple-choice questions, where the prompt lists candidate answers, but often fails on open-ended ones, where the model must produce the answer itself: a correct answer the model already samples can lose the vote to a frequent mistake, and RL then reinforces the mistake. On six ImageNet and DomainNet distribution shifts, this agreement reward lowers open-ended recognition accuracy below the unadapted model on five shifts for Qwen2.5-VL-7B and on all six for InternVL3.5-8B. We instead reward each answer by combining how often it is sampled with the score of a learned judge: a LoRA adapter on the same frozen VLM that checks each candidate against the image and question. Even without test-time RL, the source-trained judge improves answer selection over majority voting on every recognition shift and VQA benchmark for both models. As an RL reward, it beats the agreement reward on all six recognition shifts and eight VQA benchmarks, including benchmarks it never saw, by 17.4 (recognition) and 9.8 (VQA) points on average for Qwen and 15.6 and 8.2 for InternVL. A judge trained once on labeled source data is thus a more reliable reward for test-time RL than agreement, and the adapted model needs no judge at inference. Code is available at \href{https://github.com/ParsaKhani/test_time_RL}{\texttt{github.com/ParsaKhani/test\_time\_RL}}.
\end{abstract}

\vspace{-3mm}
\section{Introduction}\label{sec:intro}
\vspace{-2mm}

Vision-language models (VLMs) are trained once and then used on data their developers never saw: sketches and paintings instead of photos, book covers and scientific plots instead of natural scenes, blurry photos taken by blind users. Labels for such data are rarely available, but unlabeled questions are: a deployed model sees them every day. Test-time reinforcement learning (RL) uses them to adapt the model~\cite{zuo2025ttrl,singh2026ttrv}. For each unlabeled question, the model samples several answers, a reward scores them, and the model is updated with GRPO~\cite{shao2024deepseekmath} toward the high-reward answers. Without labels, the standard reward is \emph{consensus}: TTRL~\cite{zuo2025ttrl} rewards answers that match the majority of the samples, and TTRV~\cite{singh2026ttrv} rewards each answer by how often it was sampled. TTRV reports substantial gains on object recognition, evaluated as multiple choice, where the model selects one of four supplied labels. The success of such adaptation depends on whether agreement among the model's own answers provides a reliable reward.

An answer can win the vote even when a correct answer is already among the samples. If frequency determines the reward, the update can make the frequent mistake even more likely. This risk motivates examining open-ended recognition, where the model must generate an answer without supplied options. Prior test-time RL work for VLMs evaluates recognition only as multiple choice~\cite{singh2026ttrv}, and its follow-ups study free-form VQA~\cite{ttiq2026} or visual reasoning~\cite{ttrsd2026,timbre2026}. We instead pose recognition as open-ended answering, a harder setting where strong VLMs already struggle~\cite{ging2024ovqa,zhang2024vlmclassifier}, and show that consensus-based adaptation, which helps with supplied options, degrades both Qwen2.5-VL and InternVL3.5 here. With greedy decoding, TTRV reduces accuracy below the unadapted model on five of six distribution shifts for Qwen2.5-VL-7B~\cite{bai2025qwen25vl} (ImageNet-A drops from 26.2\% to 16.2\%) and on all six for InternVL3.5-8B~\cite{wang2025internvl35}.

During target adaptation, the model generates several candidate answers for each unlabeled question. Each generation is called a rollout. TTRV gives higher rewards to answers that appear more often among those candidates, then updates the model using those rewards. A correct answer can therefore receive less reward than a frequent mistake. On held-out ImageNet-A questions, the fraction with at least one correct answer among 32 samples falls from 61.5\% before adaptation to 36.6\% after consensus RL. The mistakes it reinforces are often broader words that are common across questions, such as \emph{bird} for a parrot or a hummingbird (\cref{fig:teaser}a). Frequency alone supplies no correctness-based reason to favor a rare correct answer over a popular wrong one. Recent work makes the consensus signal more robust, for example with rewards based on how answers depend on the image and question~\cite{ttiq2026}, self-distillation across views~\cite{ttrsd2026}, or reliability-weighted consensus over transformed inputs~\cite{timbre2026}. These signals are still derived from the model's own predictions.

To give a rare but correct answer a chance to earn higher reward, we use a \emph{judge} that checks whether the candidate is grounded in the image and answers the question: the same VLM with one extra LoRA adapter~\cite{hu2022lora}, which re-reads the image and the question with a candidate answer stated back and scores whether the answer is correct. The judge is trained once on labeled source datasets and never sees labels for the target adaptation or evaluation examples. We combine the judge score with the answer's vote share, the fraction of sampled responses that give that answer. The judge can thus give a less frequent answer a higher reward when the image supports it. In \cref{fig:teaser}a, Qwen2.5-VL answers \emph{bird} in 21 of 32 samples and the correct \emph{hummingbird} in 4; the vote rewards \emph{bird}, our reward favors \emph{hummingbird}, and after adaptation all 32 samples say \emph{hummingbird}. If the judge gives all answers to a question the same score and the frequency weight is set to one, each answer's reward equals its vote share. The judge is frozen and shares the policy's backbone, so adaptation needs no second model in memory.

For each question, we sample 32 responses from the unadapted model. Base voting returns the most frequent answer among those responses. Reranking uses the same samples and combines the judge score with vote share, allowing it to recover correct answers that lose the vote. This improves accuracy over base voting on all six recognition shifts and all eight VQA benchmarks for both models. Reranking still depends on a correct answer appearing among those samples. Can the judge's feedback instead update the model, so that it generates correct answers more often on new questions?

We therefore use the judge as the reward for test-time RL. The model samples answers to unlabeled questions, and the judge's scores help update the model toward the high-scoring ones. After adaptation, the model answers on its own, without the judge. With greedy decoding, judge-guided RL outperforms every compared test-time RL method on every recognition shift and VQA benchmark for both models, improving on TTRV by 17.4 (recognition) and 9.8 (VQA) points on average for Qwen2.5-VL and by 15.6 and 8.2 for InternVL3.5. It also goes beyond reranking: with the same 32 samples per question, voting with the adapted model beats reranking on seven of eight VQA benchmarks for both models. On OCR-VQA, only 64.6\% of questions have a correct answer among the 32 base-model samples, which caps any reranker, yet the adapted Qwen model reaches 74.3\% (\cref{sec:rerank}, \cref{fig:teaser}b).

Our main contributions are:
\begin{itemize}
\setlength{\itemsep}{1pt}\setlength{\parskip}{0pt}\setlength{\parsep}{0pt}\setlength{\topsep}{2pt}
\item \textbf{Diagnosing failures of consensus-based test-time RL in open-ended recognition.} Without answer choices, consensus RL reinforces frequent wrong answers. It lowers accuracy below the unadapted model on five of six shifts for Qwen2.5-VL and all six for InternVL3.5. On ImageNet-A, it cuts the share of questions with a correct answer among 32 samples from 61.5\% to 36.6\%.
\item \textbf{Learning transferable visual rewards for test-time RL.} We train a judge to tell correct answers from the base model's plausible errors on labeled source data. The judge is a LoRA adapter on the policy's own frozen backbone, and we combine its scores with answer frequency as the reward for test-time RL on unlabeled target questions. The judge never sees target labels and needs no second model in memory.
\item \textbf{Consistent gains, with generation beyond reranking on VQA.}  Judge-guided RL outperforms all compared test-time RL baselines across all six recognition shifts and eight VQA benchmarks for both backbones, including four VQA benchmarks excluded from judge training. The judge also improves answer selection through reranking, while adaptation can additionally generate correct answers absent from the base model's sampled candidates, as demonstrated on OCR-VQA and TabMWP.
\end{itemize}

\section{Method}\label{sec:method}

\begin{figure*}[t]
\centering
\includegraphics[width=\textwidth]{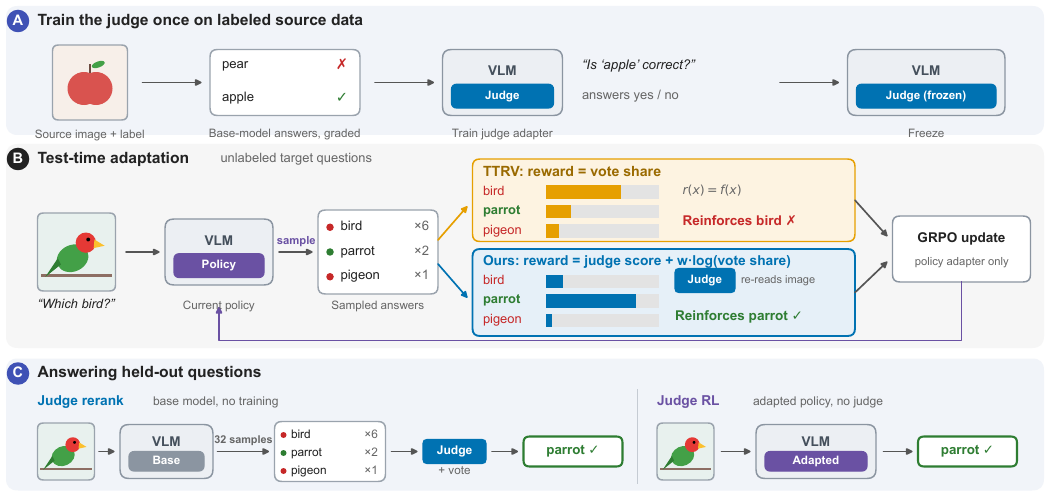}
\caption{\textbf{Overview.} (A) We train a judge once on labeled source data. The base model's answers are graded against source labels, and the judge learns to say whether a proposed answer is correct. The judge is then frozen. (B) At test time, the policy samples answers to unlabeled target questions. TTRV rewards each answer by its vote share. We add the judge's score, so a correct answer that few samples give can still earn the highest reward. GRPO updates only the policy adapter; the judge and policy are two LoRA adapters on one frozen VLM. (C) Without policy adaptation, the judge can rerank the base model's samples. After adaptation, the policy model answers on its own, without the judge.}
\label{fig:method}
\end{figure*}

\subsection{Problem formulation and overview}

Given an image $I$ and question $q$, a pretrained autoregressive VLM generates a response $y$ conditioned on the input $u=(I,q)$. We seek to improve its performance on a target distribution using an unlabeled adaptation set $\mathcal D_{\mathrm{adapt}}$. Evaluation uses a disjoint set $\mathcal D_{\mathrm{eval}}$, whose inputs are not used for adaptation.

Our approach separates supervised reward learning from target-time policy adaptation (\cref{fig:method}). We first train an answer judge $s(u,x)$ on labeled source examples to assess a proposed answer $x$. We then freeze the judge and use its scores to reward responses sampled on $\mathcal D_{\mathrm{adapt}}$. Only the policy adapter is updated during this stage. No labels from target adaptation or evaluation examples are used for training, model selection, or hyperparameter selection. Appendix~C lists source coverage and evaluation splits.

For each adaptation input, the policy generates candidate responses. We extract and canonicalize their final answers, score each distinct answer, and combine verification with candidate frequency. The resulting rewards guide GRPO updates. The adapted policy subsequently answers held-out inputs without consulting the judge.

\subsection{Learning an answer judge}

The judge evaluates a proposed answer jointly with the image and question. 
It uses the policy's pretrained backbone with a separate LoRA adapter~\cite{hu2022lora}. 
Given the prompt
\begin{quote}\small\ttfamily
<image> Question: \{q\}\\
Proposed answer: \{x\}\\
Is the proposed answer correct? Answer yes or no.
\end{quote}
let $\ell_t$ be the judge's next-token logit for token $t$. We use two disjoint sets of vocabulary tokens, $\mathcal{T}_{+}=\{\texttt{yes},\texttt{Yes}\}$ and $\mathcal{T}_{-}=\{\texttt{no},\texttt{No}\}$, to represent the verdicts \emph{correct} and \emph{incorrect}, respectively. The verification score is
\begin{equation}
s(u,x)=\log\sum_{t\in\mathcal{T}_{+}} e^{\ell_t}
             -\log\sum_{t\in\mathcal{T}_{-}} e^{\ell_t}
             =\log\frac{P(\mathcal{T}_{+}\mid u,x)}{P(\mathcal{T}_{-}\mid u,x)}.
\end{equation}
Each logit is the score before softmax for a possible first token of the judge's reply. The score is the log ratio of the total next-token probability assigned to the two verdicts. All logits come from one forward pass. For a given image and question, a higher score means the judge more strongly favors a verdict of correctness for that candidate. The next subsection shows how we incorporate this score into the adaptation reward.

\paragraph{Source supervision.} We train recognition judges on labeled ImageNet training images~\cite{russakovsky2015imagenet} and VQA judges on public VQA training sets. For each source input, we sample responses from the corresponding base model and label each distinct candidate answer as correct or incorrect using the source task's scoring rule. Sampling from the base model exposes the judge to plausible errors alongside correct answers. When a question has both, the labels tell the judge which answers to prefer for the same image and question. During target adaptation, the frozen judge uses what it learned from these source examples to score new responses on unlabeled target inputs. Appendix~B details the datasets, candidate construction, and training settings. Source labels support judge training and source-side selection, including the judge checkpoint and the vote weight $w$ in \cref{eq:reward}. For each backbone and task family, we fix the judge and reward settings before target adaptation. Policy updates use unlabeled target examples. Consensus methods use no labels, so our comparison asks whether a reward learned from other labeled data improves adaptation on a new target. The gains on benchmarks excluded from judge training (\cref{sec:results}) and the shuffled-score control (\cref{sec:ablation}) indicate that it does so through the judge's ranking of answers, not through the source labels alone.

\paragraph{Training objective.} For source candidates $\{x_j\}$ with correctness labels $c_j\in\{0,1\}$, we optimize
\begin{equation}
\mathcal{L}_{\mathrm{judge}}=\mathcal{L}_{\mathrm{BCE}}^{\mathrm{bal}}+\frac{1}{|P|}\sum_{(j,k)\in P}\log\!\left(1+e^{s(u,x_k)-s(u, x_j)}\right),
\end{equation}
where $P$ contains correct/incorrect pairs for the same question. Balanced binary cross-entropy gives correct and incorrect candidates equal total weight, while the pairwise term encourages correct answers to outrank incorrect alternatives. The pairwise term is omitted when no such pair exists. These objectives train discrimination and ranking; they do not establish calibrated correctness probabilities. 
During adaptation on unlabeled target data, we keep the judge fixed and use its scores to compute rewards for fine-tuning the VLM via its policy LoRA adapter.

\subsection{Combining verification with candidate frequency}

For an input $u$, sample $K$ responses $y_1,\ldots,y_K$ and map their extracted answers to canonical forms $x_1,\ldots,x_K$ (lowercase, punctuation removed, and number words mapped to digits). Let $\mathcal X_u$ contain the distinct answers.
Following TTRV~\cite{singh2026ttrv}, we define the empirical frequency (vote share) of $x \in \mathcal X_u$ as
\begin{equation}
f_u(x)=\frac{1}{K}\sum_{i=1}^K\mathbf{1}[x_i=x].
\end{equation}
We assign each distinct answer the combined score
\begin{equation}
z(x)=s(u,x)+w\log f_u(x),\label{eq:z}
\end{equation}
and reward response $i$ by
\begin{equation}
r_i=\frac{e^{z(x_i)}}{\sum_{x\in\mathcal X_u}e^{z(x)}}.\label{eq:reward}
\end{equation}
Normalization is over distinct answers; every response sharing an answer receives the same reward. Frequency measures how consistently the policy produces an answer, while the judge supplies a learned assessment of its correctness. A rarer answer $a$ outranks a more frequent answer $b$ precisely when
\begin{equation}
s(u,a)-s(u,b)>w\log\frac{f_u(b)}{f_u(a)}.\label{eq:rare}
\end{equation}
Thus verification can overcome a frequency disadvantage without discarding frequency information.

When $w=0$, reward depends only on judge scores. When scores are constant and $w=1$, \cref{eq:reward} reduces exactly to $r_i=f_u(x_i)$, the frequency component of TTRV. Its group-constant entropy term cancels under advantage centering. We use $w=1$ for recognition and $w=1.25$ for VQA, fixed on source data before target adaptation (Appendix~B).

\subsection{Policy optimization and deployment}

We use the combined reward in GRPO~\cite{shao2024deepseekmath}, computing a relative advantage for each sampled response,
\begin{equation}
A_i=\frac{r_i-\mathrm{mean}(r_{1:K})}{\mathrm{std}(r_{1:K})+\epsilon}.
\end{equation}
The advantages guide updates to the policy adapter, while the pretrained backbone and judge adapter remain frozen. Both reward ordering and relative spacing affect the advantages. The judge adapter is activated to score distinct answers; the policy adapter is activated for generation and policy optimization. This reuses the backbone without maintaining a second copy of the VLM. \Cref{sec:exp} reports policy optimization settings, and Appendix~B reports judge scoring overhead.

After adaptation, the policy answers without the judge, from a single response or a vote over several. To separate what the policy learns from what the judge selects, we also rerank the unadapted model's samples with the same judge and score:
\begin{equation}
\hat x_{\mathrm{rerank}}=\arg\max_{x\in\mathcal X_u}z(x).
\end{equation}
Reranking can only pick among answers the base model already samples. Adaptation instead uses the judge's feedback during training to change which answers the model generates. \cref{sec:rerank} compares the two.

\section{Experiments}\label{sec:exp}

\begin{table*}[t]
\centering\small
\setlength{\tabcolsep}{4pt}
\renewcommand{\arraystretch}{0.85}
\begin{tabular*}{\textwidth}{@{\extracolsep{\fill}}lrrrrrrr@{}}
\toprule
Target & Base & TTRV & TTRL & SCOPE & TTIQ & Judge RL & $\Delta$ \\
\midrule
\multicolumn{8}{@{}l}{\textbf{Qwen2.5-VL-7B}} \\
ImageNet-A & \underline{26.2} & 16.2 & 10.0 & 24.7 & 24.1 & \textbf{36.9} & +10.7 \\
ImageNet-R & 40.4 & 30.0 & 39.0 & 39.9 & \underline{41.6} & \textbf{58.6} & +17.0 \\
ImageNet-V2 & \underline{44.6} & 35.2 & 38.0 & 39.4 & 38.0 & \textbf{52.9} & +8.3 \\
ImageNet-Sketch & 46.5 & 39.3 & \underline{47.1} & 43.0 & 43.4 & \textbf{55.1} & +8.0 \\
DomainNet clipart & 74.3 & 71.2 & 72.9 & 65.8 & \underline{75.9} & \textbf{83.4} & +7.5 \\
DomainNet painting & 67.3 & 68.5 & 69.9 & 69.2 & \underline{71.6} & \textbf{78.0} & +6.4 \\
\midrule
\multicolumn{8}{@{}l}{\textbf{InternVL3.5-8B}} \\
ImageNet-A & \underline{11.3} & 7.3 & 7.5 & 8.7 & 8.2 & \textbf{15.7} & +4.4 \\
ImageNet-R & \underline{27.8} & 11.8 & 17.8 & 16.2 & 14.3 & \textbf{37.2} & +9.4 \\
ImageNet-V2 & \underline{30.5} & 21.9 & 27.2 & 23.9 & 19.2 & \textbf{36.4} & +5.9 \\
ImageNet-Sketch & \underline{34.4} & 26.5 & 29.6 & 27.8 & 22.2 & \textbf{39.1} & +4.7 \\
DomainNet clipart & \underline{57.9} & 54.9 & 53.5 & 55.0 & 54.0 & \textbf{69.7} & +11.8 \\
DomainNet painting & 52.5 & 48.9 & \underline{55.5} & 51.5 & 36.5 & \textbf{67.0} & +11.5 \\
\bottomrule
\end{tabular*}
\caption{Open-ended recognition accuracy (\%). Greedy decoding uses one answer per input at temperature 0. $\Delta$ is Judge RL minus the strongest greedy baseline (including Base), in percentage points. Bold marks the best greedy accuracy. Underlining marks the baseline used for $\Delta$.}
\label{tab:recognition}
\end{table*}

\subsection{Evaluation protocol}

\paragraph{Models and targets.} We evaluate Qwen2.5-VL-7B-Instruct~\cite{bai2025qwen25vl} on six recognition shifts and eight VQA benchmarks, and InternVL3.5-8B~\cite{wang2025internvl35} on the same recognition shifts and VQA benchmarks. Recognition uses a fixed open-ended prompt without supplied classes. The prompt asks for a short name of the single main object or animal, and the ground truth is an ImageNet class name. The targets are ImageNet-A~\cite{hendrycks2021nae}, ImageNet-R~\cite{hendrycks2021many}, ImageNet-V2~\cite{recht2019imagenet}, ImageNet-Sketch~\cite{wang2019sketch}, and DomainNet clipart and painting~\cite{peng2019domainnet}. VQA covers text, charts, tables, counting, and photos taken by blind users~\cite{mishra2019ocrvqa,methani2020plotqa,goyal2017vqav2,lu2023tabmwp,gurari2018vizwiz,deitke2025molmo,johnson2017clevr,masry2022chartqa}. Each run uses one frozen judge adapter for its model and task family. The recognition judges are trained on ImageNet photos. The judges for both backbones are trained on the same source data, which excludes four VQA targets (VizWiz, PixMo-Count, CLEVR, and ChartQA). Appendix~C gives the exact recognition prompt, reference labels, splits, and judge training sources.

\paragraph{Adaptation and baselines.} We adapt on 500 unlabeled questions per target for two epochs. For each input, we sample 32 responses at temperature 1, compute rewards and group-relative advantages over all 32 responses, and randomly select 16 responses for the policy gradient update. The policy LoRA uses rank 16, scaling $\alpha=32$, and learning rate $10^{-5}$, with no KL penalty. We compare with the unadapted model and with TTRV~\cite{singh2026ttrv}, TTRL~\cite{zuo2025ttrl}, SCOPE~\cite{wang2026scope}, and TTIQ~\cite{ttiq2026} where runs are available. As the official TTIQ implementation was not publicly available at the time of our experiments, we reimplemented the method following the published paper. Implementation details not specified by the authors may differ. For TTRV we use its full reward, vote share plus the entropy term. Judge RL and all RL baselines use the same adaptation questions and budget per target, with matched epochs, optimizer steps, and responses per rollout and update group. For each backbone and target, we evaluate all methods on the same held-out inputs using identical prompts, decoding, answer extraction, and scoring within each evaluation setting.

\paragraph{Decoding and scoring.} In the greedy evaluation (\cref{tab:recognition,tab:main}), each model gives one answer per held-out question at temperature 0 without the judge. This follows the decoding protocol used by TTRV and TTIQ. In the 32-sample evaluation (\cref{tab:32answers,tab:ablation}), \emph{base vote} selects the most frequent answer among 32 base-model responses at temperature 1. \emph{Judge rerank} selects from the same responses using $z(x)$ in \cref{eq:z}. \emph{RL vote} takes the most frequent answer among 32 responses generated by the adapted policy. The evaluation and adaptation splits are disjoint. We use strict class-name matching for recognition (Appendix D) and task-specific scoring after answer extraction for VQA.

\subsection{Recognition and visual question answering}\label{sec:results}

\begin{table*}[t]
\centering\small
\setlength{\tabcolsep}{4pt}
\renewcommand{\arraystretch}{0.85}
\begin{tabular*}{\textwidth}{@{\extracolsep{\fill}}lrrrrrrr@{}}
\toprule
Target & Base & TTRV & TTRL & SCOPE & TTIQ & Judge RL & $\Delta$ \\
\midrule
\multicolumn{8}{@{}l}{\textbf{Qwen2.5-VL-7B}} \\
OCR-VQA & 52.1 & \underline{64.5} & 62.6 & 56.0 & 64.3 & \textbf{74.0} & +9.5 \\
PlotQA & 75.5 & 74.9 & 74.8 & 75.9 & \underline{86.8} & \textbf{89.6} & +2.8 \\
VQAv2-count & 61.4 & 67.8 & \underline{71.3} & 69.3 & 65.0 & \textbf{74.3} & +3.0 \\
TabMWP & 67.6 & 76.7 & 79.0 & \underline{79.2} & 77.3 & \textbf{82.5} & +3.3 \\
VizWiz & 26.3 & 24.7 & 24.9 & \underline{31.6} & 25.3 & \textbf{34.8} & +3.2 \\
PixMo-Count & 62.5 & 59.4 & 50.7 & 56.5 & \underline{63.9} & \textbf{66.8} & +2.9 \\
ChartQA & 65.2 & \underline{65.6} & 64.4 & 64.3 & 63.4 & \textbf{85.9} & +20.3 \\
CLEVR & 91.5 & 93.7 & \underline{94.1} & 93.7 & 93.5 & \textbf{97.9} & +3.8 \\
\midrule
\multicolumn{8}{@{}l}{\textbf{InternVL3.5-8B}} \\
OCR-VQA & 47.3 & 45.5 & 45.6 & 45.3 & \underline{53.5} & \textbf{59.9} & +6.4 \\
PlotQA & 81.3 & 81.3 & 82.1 & 81.4 & \underline{83.7} & \textbf{87.5} & +3.8 \\
VQAv2-count & 46.3 & 64.5 & 66.5 & \underline{67.7} & 65.7 & \textbf{69.6} & +1.9 \\
TabMWP & 80.1 & \underline{85.5} & 84.9 & 85.1 & 85.1 & \textbf{89.8} & +4.3 \\
VizWiz & 25.8 & 10.2 & 22.7 & \underline{28.3} & 24.6 & \textbf{30.3} & +2.0 \\
PixMo-Count & 60.4 & 60.3 & 60.0 & 61.9 & \underline{62.2} & \textbf{65.2} & +3.0 \\
ChartQA & 85.8 & 84.9 & 85.5 & \underline{86.2} & 85.6 & \textbf{90.4} & +4.2 \\
CLEVR & \underline{92.1} & 91.1 & 90.0 & 89.5 & 91.5 & \textbf{96.4} & +4.3 \\
\bottomrule
\end{tabular*}
\caption{VQA accuracy (\%). Greedy decoding uses one answer per input at temperature 0. $\Delta$ is Judge RL minus the strongest greedy baseline (including Base), in percentage points. Bold marks the best greedy accuracy. Underlining marks the baseline used for $\Delta$.}
\label{tab:main}
\end{table*}

\noindent\textbf{Consensus RL often hurts open-ended recognition.} Without supplied options, TTRV lowers greedy accuracy below the unadapted model on five of six shifts for Qwen and on all six for InternVL (\cref{tab:recognition}); on Qwen ImageNet-A it falls from 26.2\% to 16.2\%. The other label-free baselines do not reliably help either. For InternVL, the unadapted model is the strongest baseline on five of six shifts, and only TTRL on DomainNet painting improves on it.

\noindent\textbf{Judge-guided RL is the most accurate method on every shift for both backbones.} It improves on the strongest baseline in \cref{tab:recognition}, including the unadapted model, by 6.4--17.0 percentage points for Qwen and 4.4--11.8 points for InternVL. The largest gain is on Qwen ImageNet-R, where it reaches 58.6\% compared with 41.6\% for TTIQ.

\begin{figure*}[t]
\centering
\includegraphics[width=\textwidth]{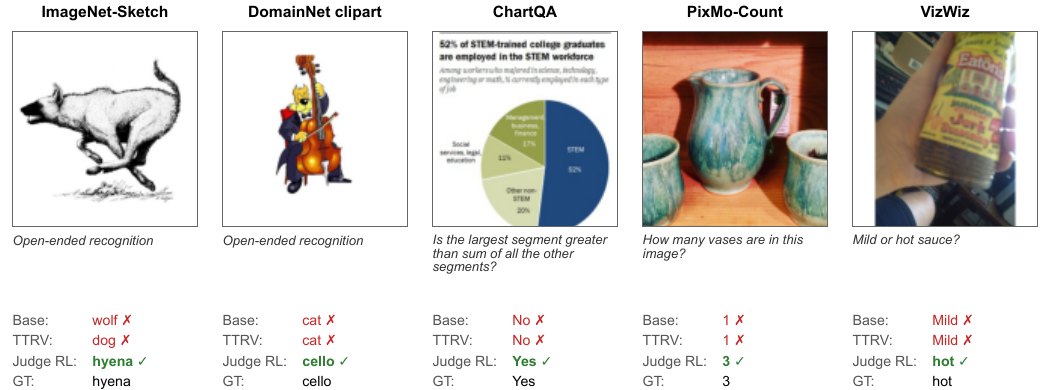}
\caption{\textbf{Greedy answers on targets outside the judge's training data} (Qwen2.5-VL-7B). Base is the unadapted model; TTRV and judge RL are adapted on unlabeled questions from each target. GT is the ground truth.}
\label{fig:qualitative}
\end{figure*}

\noindent\textbf{Judge-guided RL generalizes to benchmarks excluded from judge training.} Our judge-guided RL leads the comparison in \cref{tab:main} on all eight VQA benchmarks for both backbones. For Qwen, it improves on the strongest baseline by 2.8--20.3 points. On the four benchmarks excluded from judge training (VizWiz, PixMo-Count, ChartQA, and CLEVR), the gains are 2.9--20.3 points, with the largest on ChartQA (85.9\% vs.\ 65.6\%). For InternVL, the gains are 1.9--6.4 points, including 2.0--4.3 points on the four excluded benchmarks. Consistent improvements across all four excluded benchmarks and both backbones show that the learned reward transfers to datasets not used for judge training. \Cref{fig:qualitative} illustrates individual predictions on targets outside the judges' training data.

Appendix~D reports the paired-bootstrap analysis and answer-format audit.

\subsection{What does adaptation add to reranking?}\label{sec:rerank}

Tables~\ref{tab:recognition} and~\ref{tab:main} show that judge-guided RL improves accuracy. The judge can help in two ways: by selecting among answers sampled from the unadapted model, or by providing the reward that updates the model. To separate the two, we compare judge reranking with voting before and after adaptation, using 32 responses per evaluation question (\cref{tab:32answers}).

\noindent\textbf{The judge recovers correct answers that voting misses.} Reranking and base voting choose from the same 32 responses of the unadapted model, so their difference isolates the effect of verification without any policy update. Reranking raises mean recognition accuracy from 49.4\% to 61.0\% for Qwen and from 35.6\% to 45.1\% for InternVL, and improves on base voting on all eight VQA benchmarks for both models, by up to 15.9 points on Qwen ChartQA (InternVL in Appendix~D). Pooled over all 14 Qwen targets (15,115 questions), the base vote is wrong on 19.6\% of questions even though a correct answer is among the samples; the judge recovers 55\% of these (Appendix~D). Without any policy update, reranking already outperforms TTRV voting on every recognition shift, on six of eight Qwen VQA benchmarks, and on all eight InternVL VQA benchmarks.

\begin{table}[t]
\centering\small
\setlength{\tabcolsep}{3pt}
\renewcommand{\arraystretch}{0.9}
\begin{tabular}{@{}lrrrr@{}}
\toprule
 & \shortstack{Base\\vote} & \shortstack{TTRV\\vote} & \shortstack{Judge\\rerank} & \shortstack{Judge RL\\vote} \\
\midrule
\multicolumn{5}{@{}l}{\emph{Recognition (mean over six shifts)}} \\
Qwen & 49.4 & 43.4 & \textbf{61.0} & \textbf{61.0} \\
InternVL & 35.6 & 28.5 & \textbf{45.1} & 43.8 \\
\midrule
\multicolumn{5}{@{}l}{\emph{Qwen VQA}} \\
OCR-VQA & 52.6 & 64.5 & 59.6 & \textbf{74.3} \\
PlotQA & 73.1 & 75.1 & 77.4 & \textbf{87.3} \\
VQAv2-count & 61.2 & 68.1 & 65.9 & \textbf{72.4} \\
TabMWP & 71.7 & 76.8 & 77.1 & \textbf{80.5} \\
VizWiz & 29.1 & 24.8 & \textbf{34.0} & 33.2 \\
PixMo-Count & 61.0 & 59.4 & 62.5 & \textbf{65.4} \\
ChartQA & 65.4 & 65.4 & 81.3 & \textbf{86.3} \\
CLEVR & 91.6 & 93.9 & 97.8 & \textbf{97.9} \\
\bottomrule
\end{tabular}
\caption{Accuracy (\%) with 32 answers per input. Base vote picks the most frequent base-model answer, while judge reranking scores the same answers using the judge and vote share. TTRV and Judge RL vote on samples from their adapted models. Appendix~D gives per-shift recognition results.}
\label{tab:32answers}
\end{table}

\noindent\textbf{Adaptation improves generation beyond reranking.} A reranker can only return an answer that the base model already samples, whereas judge-guided RL changes which answers the model generates. On VQA, voting after adaptation beats reranking on seven of eight benchmarks for both models, by 5.2 points on average for Qwen and 2.6 for InternVL (Appendix~D). The largest Qwen gains are on OCR-VQA (14.7), PlotQA (9.9), VQAv2-count (6.5), and ChartQA (5.0); VizWiz slightly favors reranking for both models (by 0.8 and 0.3 points). OCR-VQA shows the difference most clearly: a correct answer appears among the 32 base samples for only 64.6\% of questions, which caps any reranker of these samples, yet the adapted policy reaches 74.3\%. About 15\% of all OCR-VQA questions and 4\% of TabMWP questions are answered correctly after adaptation even though none of their 32 base samples is correct (Appendix~D). On recognition, adaptation matches reranking for Qwen (61.0\% for both) and trails it slightly for InternVL (by 1.3 points) under 32-sample voting, possibly because the correct class is more often already among the base samples. Voting after adaptation still improves on TTRV voting by 17.6 points for Qwen and 15.3 points for InternVL.

\noindent\textbf{The judge is useful in two roles.} As a reranker, it recovers correct answers that lose the vote. As a reward, it makes correct answers more likely to be generated, especially on VQA, and the adapted model no longer needs the judge at test time.

\subsection{Ablations and alternative explanations}\label{sec:ablation}

The reward ablations use Qwen with 32-answer voting on OCR-VQA, VizWiz, and ImageNet-Sketch (\cref{tab:ablation}).

\begin{table}[t]
\centering\small
\setlength{\tabcolsep}{4pt}
\renewcommand{\arraystretch}{0.9}
\begin{tabular}{@{}lccc@{}}
\toprule
& OCR-VQA & VizWiz & Sketch \\
\midrule
Base & 52.6 & 29.1 & 45.8 \\
TTRV & 64.5 & 24.8 & 39.6 \\
\midrule
\multicolumn{4}{@{}l}{\emph{Reward}} \\
Judge scores permuted (placebo) & 51.8 & 29.5 & 35.2 \\
Sharp vote, $w=30$ & 57.0 & 26.2 & 47.5 \\
\midrule
\multicolumn{4}{@{}l}{\emph{Adaptation questions}} \\
20 & 51.9 & 29.6 & 47.2 \\
100 & 55.9 & 31.9 & 48.5 \\
500 (default reward and budget) & \textbf{74.3} & \textbf{33.2} & \textbf{55.4} \\
\bottomrule
\end{tabular}
\caption{Qwen ablation accuracy (\%) with 32-answer voting. Bold marks the best result per target.}
\label{tab:ablation}
\end{table}

\noindent\textbf{Reward alignment.} For a reward to improve accuracy, it should favor correct answers over incorrect ones. This intuition can be formalized under an idealized answer-level update (Appendix~A). Let $p$ denote the distribution over answers to a question, $c$ indicate correctness, and $r$ denote a reward with nonzero variance. Then,
\begin{equation}
\frac{d}{dt}\mathbb{E}_p[c]
= \frac{\Cov_p(r,c)}{\Std_p(r)}.
\end{equation}
Thus, expected correctness increases when reward and correctness are positively correlated. This result applies to an idealized update, not necessarily to GRPO with shared model parameters. We next examine how changing the reward affects adaptation performance.

\noindent\textbf{The gains depend on which answers the judge scores highly.} We randomly permute judge scores within each question to disrupt their alignment with candidate correctness while preserving the vote term and the distribution of judge scores. Accuracy then drops by 22.5 points on OCR-VQA and 20.2 points on ImageNet-Sketch relative to the full reward, falling below the unadapted model on both (51.8\% vs.\ 52.6\% and 35.2\% vs.\ 45.8\%). On VizWiz it stays near the unadapted model (29.5\% vs.\ 29.1\%), 3.7 points below the full reward.

\noindent\textbf{A sharper vote reward reduces the gains.} Raising the vote weight from its default to $w=30$ moves the reward toward a hard majority vote. Accuracy falls on all three benchmarks: from 74.3\% to 57.0\% on OCR-VQA, from 33.2\% to 26.2\% on VizWiz, and from 55.4\% to 47.5\% on ImageNet-Sketch. Heavily emphasizing answer frequency thus weakens the benefit of the learned judge. Removing the vote term entirely ($w=0$) also did not improve on judge + vote (Appendix~D).

\noindent\textbf{More unlabeled questions help, most on OCR-VQA.} With 20, 100, and 500 adaptation questions, accuracy rises from 51.9\% to 55.9\% to 74.3\% on OCR-VQA and from 47.2\% to 48.5\% to 55.4\% on Sketch, while VizWiz changes less (29.6\%, 31.9\%, and 33.2\%). Appendix~D gives additional reward controls.

\begin{table}[t]
\centering
\small
\setlength{\tabcolsep}{3pt}
\renewcommand{\arraystretch}{0.9}
\begin{tabular}{@{}lrrrr@{}}
\toprule
Target & Base & Source SFT & Judge RL & RL $-$ SFT \\
\midrule
ImageNet-R & 40.4 & 50.3 & \textbf{58.6} & \textbf{+8.3} \\
ImageNet-V2 & 44.6 & 50.8 & \textbf{52.9} & +2.1 \\
ImageNet-Sketch & 46.5 & 52.4 & \textbf{55.1} & +2.7 \\
\midrule
VizWiz & 26.3 & 31.4 & \textbf{34.8} & +3.4 \\
OCR-VQA & 52.1 & \textbf{75.4} & 74.0 & $-1.4$ \\
PixMo-Count & 62.5 & 61.2 & \textbf{66.8} & \textbf{+5.6} \\
\bottomrule
\end{tabular}
\caption{Qwen2.5-VL-7B greedy accuracy (\%) for source SFT and judge-guided target adaptation. The last column reports the difference in percentage points.}
\label{tab:source-sft}
\end{table}

\noindent\textbf{Does judge-guided target adaptation outperform direct fine-tuning on source labels?} Since the judge is trained on labeled source data, a simpler alternative is to fine-tune the policy directly on those labels (source SFT) and skip test-time RL. \Cref{tab:source-sft} compares the two on six Qwen targets. Judge RL is better on five: by 2.1--8.3 points on ImageNet-R, -V2, and -Sketch, 3.4 on VizWiz, and 5.6 on PixMo-Count. Source SFT is slightly better on OCR-VQA (75.4\% vs.\ 74.0\%). These results suggest that learning a reward from source labels and then adapting on unlabeled target questions can be more effective than directly fine-tuning on source labels (Appendix~D).

\noindent\textbf{A policy-specific judge outperforms a larger external judge on most targets.} Instead of training a judge, one could use an off-the-shelf external judge. We test JEV-27B-VL~\cite{autotrust2026jevvl}, a 27B vision-language decision model with a head for yes/no and multiple-choice scoring, trained on about 640{,}000 text examples~\cite{autotrust2026jev27b}. Used frozen to rerank the same Qwen base-model answers, our judge, a LoRA adapter on the 7B policy backbone, is more accurate on 12 of 14 targets, even when JEV uses its best scoring variant per target; JEV leads only on PixMo-Count and TabMWP (Appendix~D). These results show that a smaller judge trained on policy-generated answers can outperform a larger external judge, without requiring a second model in memory.

\section{Related work}\label{sec:related}

\paragraph{Learning from model predictions.} Tent~\cite{wang2021tent} minimizes prediction entropy, and test-time prompt tuning~\cite{shu2022tpt} adapts CLIP prompts to agree across views. Test-time RL also derives supervision from predictions. TTRL~\cite{zuo2025ttrl} rewards majority agreement in math, while TTRV~\cite{singh2026ttrv} uses frequency and entropy in VLMs. SCOPE~\cite{wang2026scope} weights votes by confidence, and TTIQ~\cite{ttiq2026} rewards dependence on the image and question. Related approaches use cross-view self-distillation~\cite{ttrsd2026} or reliability-weighted consensus over transformed inputs~\cite{timbre2026}. As in pseudo-label self-training~\cite{lee2013pseudo,sohn2020fixmatch}, relying on the model's predictions for supervision can reinforce its errors.

\paragraph{Adding verification to adaptation.} External feedback provides another source of supervision. RLCF~\cite{zhao2024rlcf} uses CLIP rewards for classification and for visual-projector adaptation in captioning. For text tasks, T$^3$RL~\cite{liao2026t3rl} uses executable checks to reweight votes and select math pseudo-labels, while VDS-TTT~\cite{moradi2025vdsttt} selects verifier-scored responses for fine-tuning. To verify open-ended visual answers, we build on model-based judges~\cite{zheng2023judging,xiong2025llavacritic} and generative verification via yes/no probabilities~\cite{zhang2025genverifier}. Our judge learns from labeled VQA and classification source examples. Source data may cover target task families, but exclude labels for adaptation and evaluation examples. Combining judge scores with answer frequency gives a common signal for choosing answers and learning from them.

\paragraph{Using verification for selection and learning.} Learned verifiers select among sampled solutions~\cite{cobbe2021verifiers,lightman2024verify} and can also serve as rewards for RL~\cite{wang2024mathshepherd}. We use the same judge to compare reranking a fixed set of answers with updating the policy that generates them. With constant judge scores and $w=1$, our reward reduces to frequency (\cref{eq:reward}), the quantity that self-consistency uses to select an answer~\cite{wang2023selfconsistency}. Because policies can exploit imperfect learned rewards~\cite{gao2023scaling,laidlaw2025correlated}, we state when a reward makes correct answers more likely (\cref{sec:ablation}). Under an idealized replicator flow on answer probabilities~\cite{hennes2020neural}, this condition specializes the Price equation~\cite{price1970selection} (Appendix~A).

\section{Conclusion}\label{sec:conclusion}

We identified a failure mode of consensus-based test-time RL in open-ended visual recognition: rewarding frequent answers can reinforce mistakes and make correct answers less likely to be generated. To address this, we introduced a source-trained visual judge that combines learned verification with answer frequency. Our results demonstrate two benefits. First, without updating the policy, the judge improves answer selection through reranking across all recognition shifts and VQA benchmarks for both VLMs. Second, using the judge as a reward for test-time RL improves the policy itself, outperforming all compared test-time RL baselines in greedy accuracy on all six recognition shifts and eight VQA benchmarks for both backbones, including four VQA benchmarks excluded from judge training. The adapted policy then answers independently, without requiring the judge at inference. These findings establish learned visual verification as an effective alternative to relying on answer frequency alone for test-time RL.

\section*{LLM Usage Statement}
ChatGPT (OpenAI) was used to assist with language editing, manuscript organization, and improving the clarity of technical explanations and experimental discussions. All methodological decisions, experimental results, and scientific conclusions were independently reviewed and verified by the authors.

{
    \small
    \bibliographystyle{ieeenat_fullname}
    \bibliography{refs}

@article{price1970selection,
  title   = {Selection and Covariance},
  author  = {Price, George R.},
  journal = {Nature},
  volume  = {227},
  pages   = {520--521},
  year    = {1970}
}

@inproceedings{hennes2020neural,
  title     = {Neural Replicator Dynamics: Multiagent Learning via Hedging Policy Gradients},
  author    = {Hennes, Daniel and others},
  booktitle = {AAMAS},
  year      = {2020}
}

@inproceedings{laidlaw2025correlated,
  title     = {Correlated Proxies: A New Definition and Improved Mitigation for Reward Hacking},
  author    = {Laidlaw, Cassidy and Singhal, Shivam and Dragan, Anca},
  booktitle = {ICLR},
  year      = {2025}
}

@inproceedings{singh2026ttrv,
  title     = {{TTRV}: Test-Time Reinforcement Learning for Vision Language Models},
  author    = {Singh, Akshit and Marjit, Shyam and Lin, Wei and Gavrikov, Paul and Yeung-Levy, Serena and Kuehne, Hilde and Feris, Rogerio and Doveh, Sivan and Glass, James and Mirza, M. Jehanzeb},
  booktitle = {CVPR},
  year      = {2026}
}

@article{shao2024deepseekmath,
  title   = {{DeepSeekMath}: Pushing the Limits of Mathematical Reasoning in Open Language Models},
  author  = {Shao, Zhihong and others},
  journal = {arXiv preprint arXiv:2402.03300},
  year    = {2024}
}

@article{zuo2025ttrl,
  title   = {{TTRL}: Test-Time Reinforcement Learning},
  author  = {Zuo, Yuxin and others},
  journal = {arXiv preprint arXiv:2504.16084},
  year    = {2025}
}

@inproceedings{wang2026scope,
  title     = {Beyond Majority Voting: Towards Fine-grained and More Reliable Reward Signal for Test-Time Reinforcement Learning},
  author    = {Wang, Weiqin and Wang, Yile and Chen, Kehao and Huang, Hui},
  booktitle = {Proceedings of the Annual Meeting of the Association for Computational Linguistics (ACL)},
  pages     = {37251--37265},
  year      = {2026}
}

@article{ttiq2026,
  title   = {Harnessing Image Question Dependence for Better {VLM} Test-time Reinforcement Learning},
  author  = {He, Xinrui and Li, Ting-Wei and Wang, Junting and Ai, Mengting and He, Xinyu and Tong, Hanghang and He, Jingrui},
  journal = {arXiv preprint arXiv:2609.13296},
  year    = {2026}
}

@inproceedings{wang2021tent,
  title     = {Tent: Fully Test-Time Adaptation by Entropy Minimization},
  author    = {Wang, Dequan and Shelhamer, Evan and Liu, Shaoteng and Olshausen, Bruno and Darrell, Trevor},
  booktitle = {ICLR},
  year      = {2021}
}

@inproceedings{shu2022tpt,
  title     = {Test-Time Prompt Tuning for Zero-Shot Generalization in Vision-Language Models},
  author    = {Shu, Manli and Nie, Weili and Huang, De-An and Yu, Zhiding and Goldstein, Tom and Anandkumar, Anima and Xiao, Chaowei},
  booktitle = {NeurIPS},
  year      = {2022}
}

@inproceedings{wang2023selfconsistency,
  title     = {Self-Consistency Improves Chain of Thought Reasoning in Language Models},
  author    = {Wang, Xuezhi and Wei, Jason and Schuurmans, Dale and Le, Quoc and Chi, Ed and Narang, Sharan and Chowdhery, Aakanksha and Zhou, Denny},
  booktitle = {ICLR},
  year      = {2023}
}

@article{cobbe2021verifiers,
  title   = {Training Verifiers to Solve Math Word Problems},
  author  = {Cobbe, Karl and others},
  journal = {arXiv preprint arXiv:2110.14168},
  year    = {2021}
}

@inproceedings{lightman2024verify,
  title     = {Let's Verify Step by Step},
  author    = {Lightman, Hunter and others},
  booktitle = {ICLR},
  year      = {2024}
}

@inproceedings{zhang2025genverifier,
  title     = {Generative Verifiers: Reward Modeling as Next-Token Prediction},
  author    = {Zhang, Lunjun and others},
  booktitle = {ICLR},
  year      = {2025}
}

@inproceedings{zheng2023judging,
  title     = {Judging {LLM}-as-a-Judge with {MT-Bench} and Chatbot Arena},
  author    = {Zheng, Lianmin and others},
  booktitle = {NeurIPS Datasets and Benchmarks},
  year      = {2023}
}

@inproceedings{hu2022lora,
  title     = {{LoRA}: Low-Rank Adaptation of Large Language Models},
  author    = {Hu, Edward J. and Shen, Yelong and Wallis, Phillip and Allen-Zhu, Zeyuan and Li, Yuanzhi and Wang, Shean and Wang, Lu and Chen, Weizhu},
  booktitle = {ICLR},
  year      = {2022}
}

@article{bai2025qwen25vl,
  title   = {{Qwen2.5-VL} Technical Report},
  author  = {Bai, Shuai and others},
  journal = {arXiv preprint arXiv:2502.13923},
  year    = {2025}
}

@article{wang2025internvl35,
  title   = {{InternVL3.5}: Advancing Open-Source Multimodal Models in Versatility, Reasoning, and Efficiency},
  author  = {Wang, Weiyun and others},
  journal = {arXiv preprint arXiv:2508.18265},
  year    = {2025}
}

@inproceedings{hendrycks2021nae,
  title     = {Natural Adversarial Examples},
  author    = {Hendrycks, Dan and Zhao, Kevin and Basart, Steven and Steinhardt, Jacob and Song, Dawn},
  booktitle = {CVPR},
  year      = {2021}
}

@inproceedings{hendrycks2021many,
  title     = {The Many Faces of Robustness: A Critical Analysis of Out-of-Distribution Generalization},
  author    = {Hendrycks, Dan and others},
  booktitle = {ICCV},
  year      = {2021}
}

@article{russakovsky2015imagenet,
  title     = {{ImageNet} Large Scale Visual Recognition Challenge},
  author    = {Russakovsky, Olga and Deng, Jia and Su, Hao and Krause, Jonathan and Satheesh, Sanjeev and Ma, Sean and Huang, Zhiheng and Karpathy, Andrej and Khosla, Aditya and Bernstein, Michael and Berg, Alexander C. and Fei-Fei, Li},
  journal   = {International Journal of Computer Vision},
  volume    = {115},
  number    = {3},
  pages     = {211--252},
  year      = {2015}
}

@inproceedings{recht2019imagenet,
  title     = {Do {ImageNet} Classifiers Generalize to {ImageNet}?},
  author    = {Recht, Benjamin and Roelofs, Rebecca and Schmidt, Ludwig and Shankar, Vaishaal},
  booktitle = {ICML},
  year      = {2019}
}

@inproceedings{wang2019sketch,
  title     = {Learning Robust Global Representations by Penalizing Local Predictive Power},
  author    = {Wang, Haohan and Ge, Songwei and Lipton, Zachary and Xing, Eric P.},
  booktitle = {NeurIPS},
  year      = {2019}
}

@inproceedings{peng2019domainnet,
  title     = {Moment Matching for Multi-Source Domain Adaptation},
  author    = {Peng, Xingchao and Bai, Qinxun and Xia, Xide and Huang, Zijun and Saenko, Kate and Wang, Bo},
  booktitle = {ICCV},
  year      = {2019}
}

@inproceedings{mishra2019ocrvqa,
  title     = {{OCR-VQA}: Visual Question Answering by Reading Text in Images},
  author    = {Mishra, Anand and Shekhar, Shashank and Singh, Ajeet Kumar and Chakraborty, Anirban},
  booktitle = {ICDAR},
  year      = {2019}
}

@inproceedings{methani2020plotqa,
  title     = {{PlotQA}: Reasoning over Scientific Plots},
  author    = {Methani, Nitesh and Ganguly, Pritha and Khapra, Mitesh M. and Kumar, Pratyush},
  booktitle = {WACV},
  year      = {2020}
}

@inproceedings{goyal2017vqav2,
  title     = {Making the {V} in {VQA} Matter: Elevating the Role of Image Understanding in Visual Question Answering},
  author    = {Goyal, Yash and Khot, Tejas and Summers-Stay, Douglas and Batra, Dhruv and Parikh, Devi},
  booktitle = {CVPR},
  year      = {2017}
}

@inproceedings{lu2023tabmwp,
  title     = {Dynamic Prompt Learning via Policy Gradient for Semi-structured Mathematical Reasoning},
  author    = {Lu, Pan and others},
  booktitle = {ICLR},
  year      = {2023}
}

@inproceedings{gurari2018vizwiz,
  title     = {{VizWiz} Grand Challenge: Answering Visual Questions from Blind People},
  author    = {Gurari, Danna and others},
  booktitle = {CVPR},
  year      = {2018}
}

@inproceedings{deitke2025molmo,
  title     = {Molmo and {PixMo}: Open Weights and Open Data for State-of-the-Art Vision-Language Models},
  author    = {Deitke, Matt and others},
  booktitle = {CVPR},
  year      = {2025}
}

@inproceedings{johnson2017clevr,
  title     = {{CLEVR}: A Diagnostic Dataset for Compositional Language and Elementary Visual Reasoning},
  author    = {Johnson, Justin and Hariharan, Bharath and van der Maaten, Laurens and Fei-Fei, Li and Zitnick, C. Lawrence and Girshick, Ross},
  booktitle = {CVPR},
  year      = {2017}
}

@inproceedings{lee2013pseudo,
  title     = {Pseudo-Label: The Simple and Efficient Semi-Supervised Learning Method for Deep Neural Networks},
  author    = {Lee, Dong-Hyun},
  booktitle = {ICML Workshop on Challenges in Representation Learning},
  year      = {2013}
}

@inproceedings{sohn2020fixmatch,
  title     = {{FixMatch}: Simplifying Semi-Supervised Learning with Consistency and Confidence},
  author    = {Sohn, Kihyuk and others},
  booktitle = {NeurIPS},
  year      = {2020}
}

@inproceedings{xiong2025llavacritic,
  title     = {{LLaVA-Critic}: Learning to Evaluate Multimodal Models},
  author    = {Xiong, Tianyi and others},
  booktitle = {CVPR},
  year      = {2025}
}

@inproceedings{gao2023scaling,
  title     = {Scaling Laws for Reward Model Overoptimization},
  author    = {Gao, Leo and Schulman, John and Hilton, Jacob},
  booktitle = {ICML},
  year      = {2023}
}

@article{ttrsd2026,
  title   = {{TTRSD}: Test-Time Reinforcement Learning with Self-Distillation for Vision-Language Models},
  author  = {Wang, Shuning and Wu, Zhiheng and Zhou, Xun and Cui, Chongyang and Jia, Chen and Liu, Bowen and Li, Chuanjie and Chen, Xiang and Yang, Yi and Zhang, Yumeng and Huang, Wenjie},
  journal = {arXiv preprint arXiv:2609.33414},
  year    = {2026}
}

@article{timbre2026,
  title   = {Timbre: Reliability-Aware Harmonic Consensus for Label-Free Multimodal Test-Time Reinforcement Learning},
  author  = {Lyu, Haotian and Wang, Haoyu and Wang, Ru and Ji, Feng and Li, Ziwen and Guo, Jiaxian and Gong, Mingming},
  journal = {arXiv preprint arXiv:2610.39439},
  year    = {2026}
}

@inproceedings{zhao2024rlcf,
  title     = {Test-Time Adaptation with {CLIP} Reward for Zero-Shot Generalization in Vision-Language Models},
  author    = {Zhao, Shuai and Wang, Xiaohan and Zhu, Linchao and Yang, Yi},
  booktitle = {ICLR},
  year      = {2024}
}

@inproceedings{masry2022chartqa,
  title     = {{ChartQA}: A Benchmark for Question Answering about Charts with Visual and Logical Reasoning},
  author    = {Masry, Ahmed and Long, Do Xuan and Tan, Jia Qing and Joty, Shafiq and Hoque, Enamul},
  booktitle = {Findings of ACL},
  year      = {2022}
}

@article{liao2026t3rl,
  title   = {Tool Verification for Test-Time Reinforcement Learning},
  author  = {Liao, Ruotong and R{\"o}hrich, Nikolai and Wang, Xiaohan and Zhang, Yuhui and Samadzadeh, Yasaman and Tresp, Volker and Yeung-Levy, Serena},
  journal = {arXiv preprint arXiv:2603.02203},
  year    = {2026}
}

@article{moradi2025vdsttt,
  title   = {Continuous Self-Improvement of Large Language Models by Test-Time Training with Verifier-Driven Sample Selection},
  author  = {Moradi, Mohammad Mahdi and Amer, Hossam and Mudur, Sudhir and Zhang, Weiwei and Liu, Yang and Ahmed, Walid},
  journal = {arXiv preprint arXiv:2505.19475},
  year    = {2025}
}

@misc{autotrust2026jev27b,
  author       = {{AutoTrust}},
  title        = {{JEV-27B}: Model Card and Training Details},
  howpublished = {\url{https://huggingface.co/autotrust/JEV-27B}},
  year         = {2026}
}

@misc{autotrust2026jevvl,
  author       = {{AutoTrust}},
  title        = {{autotrust/JEV-27B-VL}: A Decision Model That Learned to See Without a Single Image of Training},
  howpublished = {\url{https://huggingface.co/blog/autotrust/autotrustjev-27b-vl-a-decision-model-that-learned}},
  month        = sep,
  year         = {2026}
}

@inproceedings{ging2024ovqa,
  title     = {Open-ended {VQA} Benchmarking of Vision-Language Models by Exploiting Classification Datasets and Their Semantic Hierarchy},
  author    = {Ging, Simon and Bravo, Mar{\'i}a A. and Brox, Thomas},
  booktitle = {ICLR},
  year      = {2024}
}

@inproceedings{zhang2024vlmclassifier,
  title     = {Why are Visually-Grounded Language Models Bad at Image Classification?},
  author    = {Zhang, Yuhui and Unell, Alyssa and Wang, Xiaohan and Ghosh, Dhruba and Su, Yuchang and Schmidt, Ludwig and Yeung-Levy, Serena},
  booktitle = {NeurIPS},
  year      = {2024}
}

@inproceedings{wang2024mathshepherd,
  title     = {Math-Shepherd: Verify and Reinforce {LLMs} Step-by-step without Human Annotations},
  author    = {Wang, Peiyi and Li, Lei and Shao, Zhihong and Xu, Runxin and Dai, Damai and Li, Yifei and Chen, Deli and Wu, Yu and Sui, Zhifang},
  booktitle = {Proceedings of the Annual Meeting of the Association for Computational Linguistics (ACL)},
  pages     = {9426--9439},
  year      = {2024}
}
}

\clearpage
\appendix
\noindent\textbf{Overview.} Section~\ref{app:proofs} presents an idealized selection model relating reward alignment to answer accuracy. Section~\ref{app:judge} describes judge training, implementation details, and runtime. Section~\ref{app:eval} specifies the datasets, prompts, and training, adaptation, and evaluation splits. Section~\ref{app:additional} reports additional comparisons and diagnostics on source supervision, answer formatting, voting and reranking, adaptation gains, and reward design.

\section{Reward alignment and answer accuracy}
\label{app:proofs}

We examine how the alignment between reward and correctness affects the probability of generating a correct answer. We establish this relationship under a simplified answer-level update, which provides intuition for the reward design used in our method.

\paragraph{Setting.}
Consider a fixed question and let $p(a)$ denote the probability that the model generates answer $a$. Each answer has a correctness indicator $c(a)\in\{0,1\}$ and receives reward $r(a)$. The probability of generating a correct answer is therefore $\E_p[c]=\sum_a p(a)c(a)$.

We consider an idealized update in which the probability of each answer changes according to its reward relative to the average:
\begin{equation}
\dot p(a)
=
p(a)\frac{r(a)-\E_p[r]}{\Std_p(r)},
\label{eq:selection_flow}
\end{equation}
where $\dot p(a)$ denotes the rate of change of $p(a)$ during adaptation and $\Std_p(r)>0$. Since the expected standardized reward is zero, this update preserves $\sum_a p(a)=1$. The reward may itself depend on the current answer distribution, as in frequency-based rewards.

\paragraph{Reward-alignment identity.}
Differentiating the probability of a correct answer under this update gives
\begin{equation}
\begin{aligned}
\frac{d}{dt}\E_p[c]
&=\sum_a c(a)\dot p(a)\\
&=\frac{\E_p[cr]-\E_p[c]\E_p[r]}{\Std_p(r)}\\
&=\frac{\Cov_p(r,c)}{\Std_p(r)}.
\end{aligned}
\label{eq:reward_alignment}
\end{equation}
Thus, under the assumed update, the probability of generating a correct answer increases when reward and correctness are positively correlated. In particular, when both correct and incorrect answers have nonzero probability, positive covariance means that correct answers receive higher reward on average than incorrect ones.

\paragraph{Implications for reward design.}
This identity helps explain the failure of consensus-based rewards. \textbf{When an incorrect answer dominates the model's predictions, rewarding answers primarily by their frequency can produce negative covariance with correctness, shifting probability away from correct answers.} Our method addresses this issue by incorporating a learned judge that evaluates candidate answers against the image and question. By assigning higher rewards to answers supported by the visual evidence, the judge can improve reward--correctness alignment, including when a correct answer receives fewer votes.

The derivation characterizes an idealized answer-level selection process rather than the exact dynamics of GRPO, which uses sampled advantages and token-level updates with shared model parameters. Ordinary policy gradients need not follow this flow~\cite{hennes2020neural}. The derived identity is a specialization of the Price equation~\cite{price1970selection} and provides a theoretical interpretation of the reward-alignment mechanism.

\section{Judge training and implementation details}\label{app:judge}

\paragraph{Training data.} We train each judge adapter once by supervised fine-tuning on labeled source answer-verification examples; no RL is used to train the judge. Each example contains an image, question, and candidate answer labeled correct or incorrect, and the judge learns to predict Yes or No. For each backbone (Qwen2.5-VL and InternVL3.5), we use an ImageNet judge for recognition and a VQA judge for visual question answering. Both backbones use the same source datasets. The two judge adapters share their model's backbone. Each run uses only the judge adapter for its task family.
\begin{itemize}
\item \emph{VQA judges.} Train splits of 15 public tasks: VQAv2, TextVQA, GQA, A-OKVQA (open-ended and multiple-choice), AI2D, ScienceQA, DocVQA, OCR-VQA, DVQA, PlotQA, TallyQA, FigureQA, TabMWP and MapWise. We use up to 2,000 questions per task and remove any question whose image is a near-duplicate of a target image (perceptual hash). For each question, the candidates are the distinct answers among 16 base-model samples, each labeled right or wrong with the task's own metric. We add one short generic answer per question (\eg ``yes'', ``0'', ``unanswerable'') as an extra candidate, labeled by the same metric, so that the judge does not reward degenerate one-word answers that RL could drift toward. Each question keeps at most 8 candidates, with up to 4 correct.
\item \emph{Recognition judges.} 32,000 ImageNet training images from 800 classes, with the fixed question ``What is the single main object or animal in this image?''. Candidates are the base model's sampled answers, labeled by lenient class-name matching.
\end{itemize}

For each source question we sample K rollouts at T=1.0 (top-p 1, no top-k, 48 new tokens), group them by a canonical answer key, drop empty answers, and keep all distinct answers, at most 8 per question (half correct and half wrong when capped). We add one synthetic generic answer per question, scored with the task metric, and never insert the gold answer. K=16 for VQA judges and K=32 for recognition judges. Source data, sampling, deduplication and retention are the same for Qwen and InternVL.\paragraph{Training objective.} We minimize balanced binary cross-entropy plus a within-question pairwise logistic ranking loss. Balanced BCE gives correct and incorrect candidates equal total weight; the ranking term encourages each correct candidate to score above incorrect candidates and is omitted when no such pair exists.\paragraph{Optimization.} For both Qwen and InternVL, the recognition and VQA judge adapters use rank-16 LoRA on all attention and MLP projections of the language model; the vision encoder is untouched. Each judge is trained for 6,000 steps with AdamW and learning rate $10^{-4}$, using one source question or image per step and uniform task sampling for VQA. The checkpoint is chosen on held-out source examples by how much selecting the top-scored answer beats majority vote.

\paragraph{Answer verification design.} The judge re-reads a proposed answer alongside the image and question. The positive token set contains the vocabulary tokens for \texttt{yes} and \texttt{Yes}, and the negative set contains those for \texttt{no} and \texttt{No}. We sum the next-token probabilities within each set before computing the log ratio. Sharing base weights permits adapter switching during RL, while sampling source candidates from the base policy exposes the judge to the policy's near-miss errors.

\paragraph{Vote weight.} We selected $w$ once on held-out source validation data by searching $w\in[0,6]$ and maximizing judge-reranking accuracy. The VQA search used 4,413 questions spanning the 15 source tasks; the recognition search used held-out ImageNet source examples. This gave $w=1$ for recognition and $w=1.25$ for VQA. Each value was shared across all corresponding targets, fixed before target adaptation, and not tuned per target.
\paragraph{Runtime.} The frozen judge adapter is activated only to score distinct answers, and the policy adapter is used for generation and optimization. Inside the RL process, scores match the offline judge ($r=.9999$). The additional computation is one forward pass per distinct answer per question. In our runs, judge-reward RL took 4--9\% longer than matching consensus RL.

\paragraph{TTIQ.} No public implementation is available, so we reimplement TTIQ from its paper \cite{ttiq2026}. For each rollout, we compute per-token log-probability drops after replacing the image with zeros and after masking the question tokens, convert both signals to within-group ranks, and use $\min\{r_{\mathrm{image}}, r_{\mathrm{question}}, r_{\mathrm{confidence}}\}$ as the reward, with $\lambda=0.5$ fixed across targets and TTIQ's token-level credit weights. This requires three additional forward passes per rollout. TTIQ uses the same adaptation questions, 32-response budget, number of epochs, LoRA configuration, and learning rate ($10^{-5}$) as the other methods, and is evaluated with greedy decoding.

\section{Evaluation splits}\label{app:eval}

\begin{table}[t]
\centering\footnotesize
\setlength{\tabcolsep}{3pt}
\begin{tabular}{@{}llrc@{}}
\toprule
Target & Images & $n$ & Judge training \\
\midrule
ImageNet-A~\cite{hendrycks2021nae} & hard photos & 1000 & ImageNet train \\
ImageNet-R~\cite{hendrycks2021many} & renditions & 1000 & ImageNet train \\
ImageNet-V2~\cite{recht2019imagenet} & new photos & 1000 & ImageNet train \\
ImageNet-Sketch~\cite{wang2019sketch} & sketches & 1000 & ImageNet train \\
DomainNet clipart~\cite{peng2019domainnet} & clipart & 1000 & ImageNet train \\
DomainNet painting~\cite{peng2019domainnet} & paintings & 1000 & ImageNet train \\
\midrule
OCR-VQA~\cite{mishra2019ocrvqa} & book covers & 989 & Target train \\
PlotQA~\cite{methani2020plotqa} & plots & 959 & Target train \\
VQAv2-count~\cite{goyal2017vqav2} & COCO photos & 1500 & Target train \\
TabMWP~\cite{lu2023tabmwp} & tables & 1500 & Target train \\
VizWiz~\cite{gurari2018vizwiz} & blind users' photos & 1000 & -- \\
PixMo-Count~\cite{deitke2025molmo} & web photos & 667 & -- \\
CLEVR~\cite{johnson2017clevr} & 3D renders & 1500 & -- \\
ChartQA~\cite{masry2022chartqa} & charts & 1000 & -- \\
\bottomrule
\end{tabular}
\caption{Evaluation targets and judge training splits. The recognition judge was trained on photographs from the ImageNet training split. For VQA, \emph{Target train} means the judge was trained on that benchmark's training split. A dash marks a benchmark excluded from judge training. Training images are disjoint from adaptation and evaluation images.}
\label{tab:targets}
\end{table}

\Cref{tab:targets} lists the targets and evaluation sizes. For recognition, we pair each image with one fixed prompt that lists no classes: ``Look at the image and identify the single main object or animal in it. Give a SHORT answer (a common name, at most a few words) -- do not describe the whole scene. End your response with a new line that reads exactly: Final answer: \textless{}your short answer\textgreater{}''. The 500 adaptation and 1,000 evaluation images are fixed random draws, shared by both models. The ground truth is the first WordNet lemma of the ImageNet class: ImageNet-A and ImageNet-R use their 200 classes, and ImageNet-V2 and ImageNet-Sketch the 1,000 ImageNet classes. For DomainNet clipart and painting, we use only the classes that also exist in ImageNet.

For VQA, we use VQAv2 questions beginning with \emph{how many} and CLEVR CoGenT validation-A counting and count-comparison questions. For every target, the adaptation and evaluation splits are disjoint, and judge training excludes both. Each full-budget adaptation uses 126 steps with eight questions per step. Each reported adaptation result uses one fixed random seed per model--target--method configuration.

Adaptation and evaluation splits are disjoint at the image level, and adaptation questions are unlabeled during RL. Evaluation examples are sampled from the designated held-out splits of each dataset.

\section{Additional results and diagnostics}
\label{app:additional}

\subsection{Direct supervised fine-tuning on source data}

We fine-tune Qwen2.5-VL-7B directly on labeled source examples. The recognition model uses the full set of 32,000 ImageNet source images from 800 classes for 1,000 optimization steps. The VQA model uses the labeled VQA source tasks listed in Appendix~\ref{app:judge} for 1,133 optimization steps. Both models are evaluated with greedy decoding, using strict answer extraction on 1,000 examples per recognition target and lenient extraction on VQA, matching the main-text evaluation.

\subsection{Paired accuracy and answer-format checks}

\paragraph{Paired accuracy.} We resample evaluation-question indices within each target while keeping Base and Judge RL scores paired. \Cref{tab:paired-bootstrap} gives nominal, unadjusted 95\% intervals from $B=5{,}000$ paired-bootstrap replicates for Judge RL minus Base; all 14 exclude zero. Thus, Judge RL shows a consistent positive improvement over Base at the individual-target level under paired resampling; because the intervals are unadjusted, they do not provide a family-wise error guarantee across targets.

\begin{table}[t]
\centering\small
\setlength{\tabcolsep}{4pt}
\renewcommand{\arraystretch}{0.86}
\begin{tabular}{@{}lrr@{}}
\toprule
Target & $\Delta$ (pts) & 95\% CI \\
\midrule
ImageNet-A & +10.7 & $[+8.4,+13.1]$ \\
ImageNet-R & +18.2 & $[+15.8,+20.8]$ \\
ImageNet-V2 & +8.3 & $[+5.9,+10.9]$ \\
ImageNet-Sketch & +8.6 & $[+6.3,+11.0]$ \\
DomainNet clipart & +9.1 & $[+7.0,+11.4]$ \\
DomainNet painting & +10.7 & $[+8.6,+12.9]$ \\
\midrule
OCR-VQA & +21.9 & $[+19.1,+24.8]$ \\
PlotQA & +14.1 & $[+11.4,+16.8]$ \\
VQAv2-count & +12.9 & $[+10.3,+15.6]$ \\
TabMWP & +14.9 & $[+12.5,+17.4]$ \\
VizWiz & +8.5 & $[+5.9,+11.1]$ \\
PixMo-Count & +4.3 & $[+1.3,+7.2]$ \\
ChartQA & +20.7 & $[+18.0,+23.4]$ \\
CLEVR & +6.4 & $[+5.0,+7.9]$ \\
\bottomrule
\end{tabular}
\caption{Paired-bootstrap 95\% confidence intervals for Qwen Judge RL minus Base, in percentage points. Questions are resampled in paired form.}
\label{tab:paired-bootstrap}
\end{table}

\paragraph{Answer format.} We conducted an exhaustive answer-format audit across all 14 evaluation targets, examining every case in which Judge RL corrected a Base prediction. We distinguished genuine answer-content improvements from formatting-only corrections. At most 5\% of corrections were formatting-only, with none observed on OCR-VQA or ImageNet-Sketch.

\paragraph{Recognition scoring.} We use the same recognition scorer for all methods. We extract the text following ``Final answer:'' (or the last line if the marker is absent), lowercase it, remove punctuation and the words \{a, an, the, small, large, young, wild, domestic\}, and ignore spaces during matching. A prediction is counted as correct if the complete reference class name appears as a contiguous substring of the processed prediction. Unlike a more lenient rule that accepts the final word of a multiword class name, our evaluation requires the full reference class name. This is the class-name matching criterion referred to as ``strict'' in the main text; it is not exact string equality. Substring matching permits plurals and compounds, such as ``ducks'' for ``duck''. The same criterion is applied to Base, all test-time RL baselines, and Judge RL.

\begin{table*}[t]
\centering\small
\setlength{\tabcolsep}{4pt}
\begin{tabular*}{\textwidth}{@{\extracolsep{\fill}}lrrrrrr@{}}
\toprule
  &   &   & \multicolumn{4}{c}{JEV-27B-VL} \\
\cmidrule(l){4-7}
Target & Base vote & Judge rerank & Yes & Yes+vote & Choice & Choice+vote \\
\midrule
\multicolumn{7}{@{}l}{\emph{Recognition}} \\
ImageNet-A & 26.2 & \textbf{36.5} & 22.0 & 24.7 & 27.3 & 28.3 \\
ImageNet-R & 40.2 & \textbf{54.9} & 33.0 & 38.7 & 49.0 & 46.3 \\
ImageNet-V2 & 44.0 & \textbf{57.5} & 38.3 & 44.0 & 49.0 & 47.3 \\
ImageNet-Sketch & 45.8 & \textbf{57.2} & 49.3 & 53.3 & 53.3 & 53.7 \\
DomainNet clipart & 73.7 & \textbf{83.2} & 66.3 & 69.0 & 70.7 & 70.7 \\
DomainNet painting & 66.7 & \textbf{76.7} & 63.7 & 68.3 & 71.3 & 72.0 \\
\midrule
\multicolumn{7}{@{}l}{\emph{VQA}} \\
OCR-VQA & 52.6 & \textbf{59.6} & 52.4 & 55.7 & 54.7 & 56.4 \\
PlotQA & 73.1 & \textbf{77.4} & 65.3 & 75.0 & 70.1 & 75.3 \\
VQAv2-count & 61.2 & \textbf{65.9} & 36.4 & 64.6 & 56.3 & 65.6 \\
TabMWP & 71.7 & 77.1 & 62.7 & \textbf{79.0} & 60.0 & 76.0 \\
VizWiz & 29.1 & \textbf{34.0} & 20.3 & 28.3 & 16.3 & 26.3 \\
PixMo-Count & 61.0 & 62.5 & \textbf{74.0} & \textbf{74.0} & 73.0 & 71.7 \\
ChartQA & 65.4 & \textbf{81.3} & 63.0 & 66.7 & 57.7 & 65.7 \\
CLEVR & 91.6 & \textbf{97.8} & 83.7 & 90.0 & 84.3 & 92.3 \\
\bottomrule
\end{tabular*}
\caption{Reranking accuracy (\%) of our Qwen2.5-VL policy-specific verifier and JEV-27B-VL under the same evaluation protocol and held-out target splits used in the main text. Base vote picks the most frequent answer; Judge rerank combines our judge score with log vote share. Yes scores each distinct answer independently. Choice scores the same distinct candidate answers jointly in a single prompt; all judge variants use the same candidate set and candidate count. The vote variants add a log vote-share term. Bold marks the best selector per target.}
\label{tab:jev_all}
\end{table*}

\subsection{Voting, single-answer accuracy, and reranker variants}
\Cref{tab:sampled-qwen-rec,tab:sampled-internvl-rec,tab:sampled-internvl-vqa} report the 32-sample results: reranking improves over Base on every target.

\begin{table}[t]
\centering\small
\setlength{\tabcolsep}{2pt}
\begin{tabular}{@{}lrrrr@{}}
\toprule
Target & \shortstack{Base\\vote} & \shortstack{TTRV\\vote} & \shortstack{Judge\\rerank} & \shortstack{Judge RL\\vote} \\
\midrule
ImageNet-A & 26.2 & 16.6 & 36.5 & \textbf{37.2} \\
ImageNet-R & 40.2 & 30.3 & 54.9 & \textbf{58.6} \\
ImageNet-V2 & 44.0 & 34.9 & \textbf{57.5} & 53.3 \\
ImageNet-Sketch & 45.8 & 39.6 & \textbf{57.2} & 55.4 \\
DomainNet clipart & 73.7 & 70.9 & 83.2 & \textbf{83.6} \\
DomainNet painting & 66.7 & 68.1 & 76.7 & \textbf{77.8} \\
\bottomrule
\end{tabular}
\caption{Qwen2.5-VL-7B recognition accuracy (\%). Base vote and judge reranking share 32 base-model answers. TTRV and Judge RL vote on 32 answers from their own policies. Best per row is bold.}
\label{tab:sampled-qwen-rec}
\end{table}

\begin{table}[t]
\centering\small
\setlength{\tabcolsep}{2pt}
\begin{tabular}{@{}lrrrr@{}}
\toprule
Target & \shortstack{Base\\vote} & \shortstack{TTRV\\vote} & \shortstack{Judge\\rerank} & \shortstack{Judge RL\\vote} \\
\midrule
ImageNet-A & 11.2 & 7.1 & \textbf{17.9} & 16.0 \\
ImageNet-R & 27.5 & 12.0 & 37.1 & \textbf{37.6} \\
ImageNet-V2 & 30.0 & 21.8 & \textbf{40.9} & 36.6 \\
ImageNet-Sketch & 34.5 & 26.5 & \textbf{43.5} & 35.8 \\
DomainNet clipart & 57.0 & 54.8 & 66.2 & \textbf{69.6} \\
DomainNet painting & 53.4 & 48.5 & 65.0 & \textbf{67.1} \\
\bottomrule
\end{tabular}
\caption{InternVL3.5-8B recognition accuracy (\%). Base vote and judge reranking share 32 base-model answers. TTRV and Judge RL vote on 32 answers from their own policies. Best per row is bold.}
\label{tab:sampled-internvl-rec}
\end{table}

\begin{table}[t]
\centering\small
\setlength{\tabcolsep}{2pt}
\begin{tabular}{@{}lrrrr@{}}
\toprule
Target
& \shortstack{Base\\vote}
& \shortstack{TTRV\\vote}
& \shortstack{Judge\\rerank}
& \shortstack{Judge RL\\vote} \\
\midrule
OCR-VQA
& 47.7 & 45.9 & 54.1 & \textbf{60.5} \\
PlotQA
& 80.8 & 81.5 & 84.6
& \textbf{87.1} \\
VQAv2-count
& 45.2 & 64.7 & 66.6 & \textbf{69.3} \\
TabMWP
& 83.8 & 85.9
& 86.9 & \textbf{88.2} \\
VizWiz
& 28.1 & 10.2
& \textbf{31.2} & 30.9 \\
PixMo-Count
& 60.9 & 60.0
& 61.8 & \textbf{64.6} \\
ChartQA
& 86.1 & 85.0
& 87.7 & \textbf{90.8} \\
CLEVR
& 91.9 & 90.9 & 94.0 & \textbf{96.5} \\
\bottomrule
\end{tabular}
\caption{InternVL3.5-8B VQA accuracy (\%). Base vote and judge reranking share 32 base-model answers. TTRV and Judge RL vote on 32 answers from their own policies. All columns use lenient answer extraction. Best per row is bold.}
\label{tab:sampled-internvl-vqa}
\end{table}

\paragraph{JEV comparison and decoding.} \Cref{tab:jev_all} shows that our verifier is best on 12/14 targets; JEV leads only on TabMWP and PixMo-Count. Qwen judge RL reaches 52.9\%, 55.1\%, 83.4\%, and 78.0\% with greedy decoding on V2, Sketch, clipart, and painting, respectively. These differ from its 32-sample votes by at most 0.4 percentage points. Greedy decoding uses $T=0$.

\paragraph{Consensus failures.} We pool the 15,115 Qwen evaluation questions across six recognition and eight VQA targets. We count a failure when the base vote is incorrect and at least one of the same 32 base-model responses is correct. These failures account for 19.6\% of all questions. A failure is recovered when the judge-based reranker selects a correct answer from that pool, which it does in 55\% of these cases.

\subsection{Where adaptation improves on reranking}\begin{table}[!ht]\centering\small\setlength{\tabcolsep}{4pt}\begin{tabular}{@{}lrrr@{}}\toprule{} Target & Judge rerank & Judge RL vote & $\Delta$ \\\midrule{} OCR-VQA & 59.6 & 74.3 & +14.7 \\PlotQA & 77.4 & 87.3 & +9.9 \\VQAv2-count & 65.9 & 72.4 & +6.5 \\TabMWP & 77.1 & 80.5 & +3.4 \\VizWiz & 34.0 & 33.2 & -0.8 \\PixMo-Count & 62.5 & 65.4 & +2.9 \\ChartQA & 81.3 & 86.3 & +5.0 \\CLEVR & 97.8 & 97.9 & +0.1 \\\bottomrule\end{tabular}\caption{Qwen VQA accuracy (\%) with 32 responses per input. Judge reranking selects among base-model responses; Judge RL votes on responses from the adapted policy. $\Delta$ is Judge RL minus reranking.}\label{tab:rerank-vs-rl}\end{table}\Cref{tab:rerank-vs-rl} shows that Judge RL improves over judge reranking on seven of eight VQA targets, with the largest gains on OCR-VQA and PlotQA. On OCR-VQA, about 15\% of questions are answered correctly after RL even though all 32 base responses are incorrect. The corresponding fraction is 4\% on TabMWP. On VQAv2-count and PixMo-Count, a correct minority answer missed by the reranker becomes the adapted model's most frequent answer on 7\% and 5\% of questions, respectively. On 2--4\% of questions on the six analyzed targets other than OCR-VQA, the base vote is correct, the reranker overrides it, and the adapted vote preserves it. 

\subsection{Additional reward controls}
\begin{table}[!ht]
\centering\small
\setlength{\tabcolsep}{3pt}
\begin{tabular}{@{}lrrrr@{}}
\toprule
Target & Base & TTRV & \shortstack{Judge + vote\\default $w$} & \shortstack{Judge only\\$w=0$} \\
\midrule
ImageNet-A & 26.2 & 16.2 & \textbf{36.9} & 29.4 \\
ImageNet-Sketch & 46.5 & 39.3 & \textbf{55.1} & 51.4 \\
OCR-VQA & 52.1 & 64.5 & \textbf{74.0} & 71.9 \\
VizWiz & 26.3 & 24.7 & \textbf{34.8} & 30.3 \\
\bottomrule
\end{tabular}
\caption{Effect of removing the vote term (accuracy, \%). The default uses $w=1$ for recognition and $w=1.25$ for VQA. All results use greedy decoding. Best per row is bold.}
\label{tab:w0}
\end{table}
\Cref{tab:w0} shows that the judge-only reward remains above Base and TTRV on all four targets, while adding the vote term improves accuracy by 7.5, 3.7, 2.1, and 4.5 points, respectively. Shuffling the judge scores weakens reranking: the mean gain over base vote falls from 5.6 points without shuffling to 0.4 with 25\% of scores shuffled and $-18.0$ with all scores shuffled.
\FloatBarrier

\end{document}